\documentclass[runningheads]{llncs}
\usepackage{graphicx} 
\usepackage{amssymb} 
\usepackage{graphicx}  
\usepackage{subcaption} 
\usepackage{caption}
\usepackage{tabu} 
\usepackage{booktabs} 
\usepackage{mathtools}  
\usepackage{multirow} 
\usepackage{xcolor} 
\usepackage{hyperref}
\usepackage{enumitem}
\setlist{nosep,after=\vspace{\baselineskip}}

\begin{document}
\title{Morphological Networks for Image De-raining}
\author{Ranjan Mondal\inst{1} \and
Pulak	Purkait \inst{2}\and
Sanchayan	Santra\inst{1}\and
Bhabatosh	Chanda \inst{1} }
\institute{Indian Statistical Institute, Kolkata, India  \\
\email{\{ranjan15\_r,sanchayan\_r,chanda\}@isical.ac.in} \and
The University of Adelaide, Adelaide, Australia \\
\email{pulak.isi@gmail.com}}

\authorrunning{Ranjan et al.}

%

\maketitle 

\begin{abstract}
Mathematical morphological methods have successfully been applied to filter out (emphasize or remove) different structures of an image. However, it is argued that these methods could be suitable for the task only if the type and order of the filter(s) as well as the shape and size of operator kernel are designed properly. Thus the existing filtering operators are problem (instance) specific and are designed by the domain experts. In this work we propose a morphological network that emulates classical morphological filtering consisting of a series of erosion and dilation operators with trainable structuring elements. We evaluate the proposed network for image de-raining task where the SSIM and mean absolute error (MAE) loss corresponding to predicted and ground-truth clean image is back-propagated through the network to train the structuring elements. We observe that a single morphological network can de-rain an image with any arbitrary shaped rain-droplets and achieves similar performance with the contemporary CNNs for this task with a fraction of trainable parameters (network size). The proposed morphological network(MorphoN) is not designed specifically for de-raining and can readily be applied to similar filtering / noise cleaning tasks. The source code can be found here  \href{https://github.com/ranjanZ/2D-Morphological-Network}{https://github.com/ranjanZ/2D-Morphological-Network}

\keywords{Mathematical Morphology  \and Optimization \and Morphological Network \and Image Filtering}
\end{abstract}
   
\section{Introduction} 

Morphological Image processing with hand-crafted filtering operators has been applied successfully to solve many problems like image segmentation (\cite{wdowiak2015hourglass,perret2015evaluation}), object shape detection and noise filtering. Due to rich mathematical foundation, image analysis by morphological operators is found to be very effective and popular. The basic building block operations are \textit{dilation} and \textit{erosion}, which are defined in terms of a structuring element(SE). Many problem can be solved by choosing shape and size of the structuring element intelligently~\cite{vincent1993morphological}. Finding customized / tailored size and shape of the structuring elements and also the order in which erosion and dilation operations are to be applied still remain a huge challenge. Furthermore, the design could be data dependent, \emph{i.e.,} the expert might have to design the operator depending on the problem instances. For example, for de-raining task one needs to design different filtering operator for different rain pattern.  

In this work we utilize a network architecture that consists of trainable morphological operators to de-rain the rainy images irrespective of the rain pattern. 
\begin{figure}[t]
    \centering
    
    \begin{subfigure}[h!]{0.30\linewidth}
        \includegraphics[width=\linewidth]{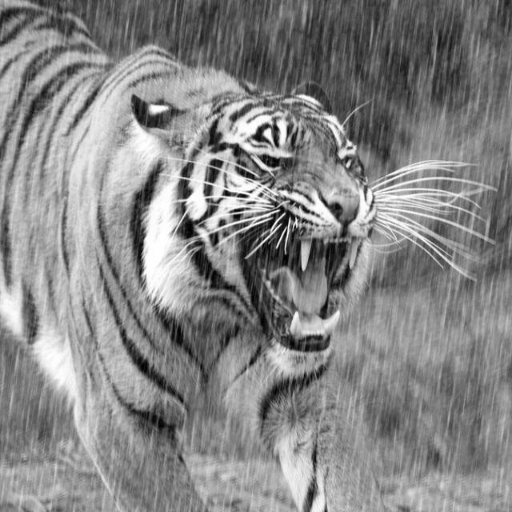}
        \caption{Input}
        \label{fig:input1}
    \end{subfigure}
    \begin{subfigure}[h!]{0.30\linewidth}
        \includegraphics[width=\linewidth]{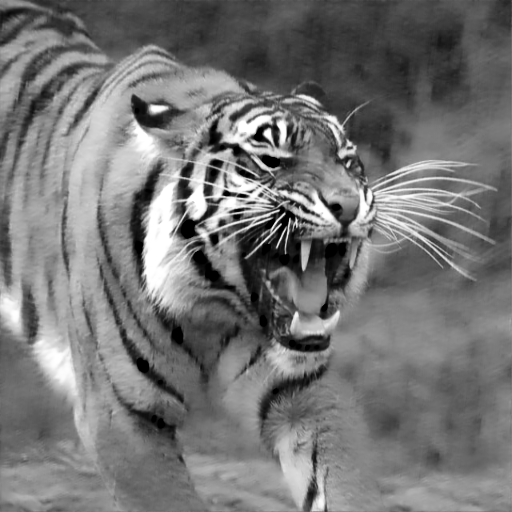}
         \caption{CNN ({\#para 6M})}
        \label{fig:CNN1}
    \end{subfigure}
    \begin{subfigure}[h!]{0.30\linewidth}
        \includegraphics[width=\linewidth]{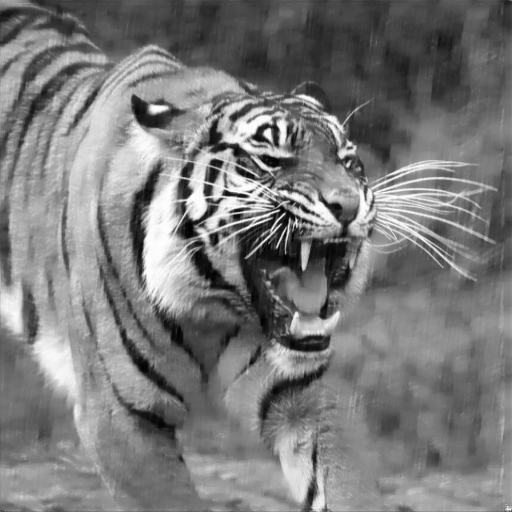}
        \caption{\scalebox{0.91}{MorphoN({\#para 2K})}}
        \label{fig:output1}
    \end{subfigure}
    \begin{subfigure}[h!]{0.30\linewidth}
        \includegraphics[width=\linewidth]{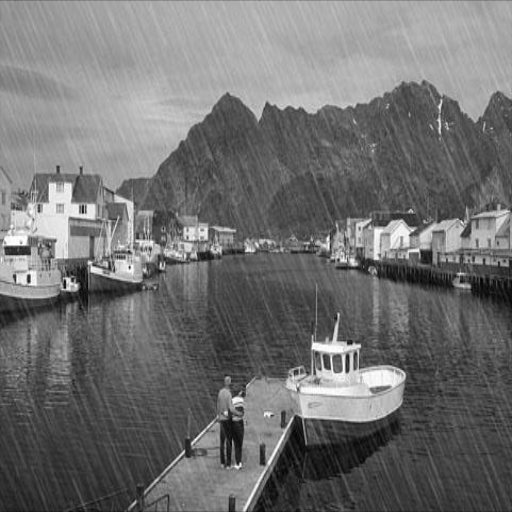}
        \caption{Input}
        \label{fig:input2}
    \end{subfigure}
    \begin{subfigure}[h!]{0.30\linewidth}
        \includegraphics[width=\linewidth]{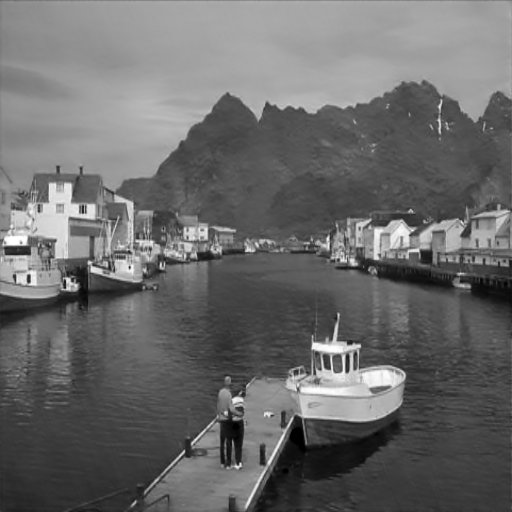}
        \caption{CNN({\#para 6M})}
        \label{fig:CNN2}
    \end{subfigure}
    \begin{subfigure}[h!]{0.30\linewidth}
        \includegraphics[width=\linewidth]{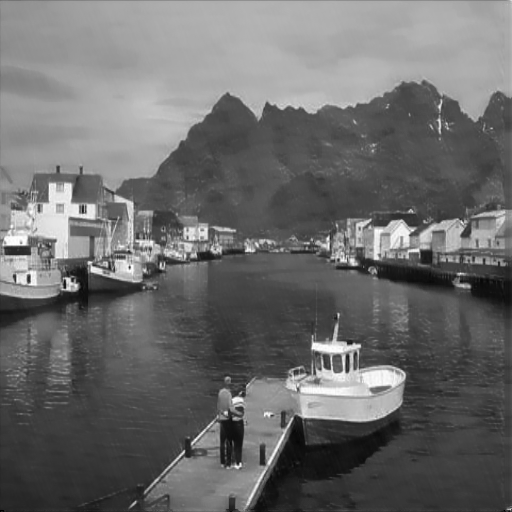}
        
        \caption{\scalebox{0.91}{MorphoN({\#para 2K})}}
        
        \label{fig:output2}
    \end{subfigure}
    \caption{ Some examples of the results by the proposed Morphological Network. Note that the same network was used to clean different types and amounts of rain---(a) input rainy image with vertical structures and (d) input rainy image with sliding structures. The proposed network emulates standard 2D morphological operations where the structuring elements are trained using the back propagation. The size of our network is drastically smaller than the conventional CNNs and capable of producing very high quality results ((c) and (f) vs (b) and (e)). More results can be found in the experiment section. }
    \label{dp_paper}
    
\end{figure}

\subsection{Motivation and Contributions}\label{sec:motivation} 

The recent developments of Convolution Neural Networks (CNN) have unveiled a huge success in image processing and computer vision tasks. A number of examples could be found in~\cite{goodfellow2016deep}. To mention a few, CNNs are very popular in solving problems like object detection~\cite{ren2015faster}, image dehazing~\cite{mondal2018image} and image segmentation~\cite{chen2018deeplab}. A CNN consists of an input layer, an output layer, and multiple hidden layers. An image input passes through the stack of hidden layers of the trained network and produces the desired  output. The hidden layers of a CNN typically consist of convolutional layers and optionally---pooling layers, fully connected layers and normalization layers. The convolutional layers apply a convolution operation with a trained mask on the input and pass the result to the next layer. 

Inspired by the success of convolutional networks and similarity between the convolution and morphological dilation and erosion (both are neighbourhood operator with respect to some kernel), we propose morphological layers by replacing convolution operators by $\max$ or $\min$ operator that yields morphological networks.
The network consists of a sequence of dilation-erosion operators for which the structuring elements are trained using back-propagation for a particular task, 
very similar to the way we train the weights of a convolutional layers of CNNs. An example of results after learning the structuring elements is displayed in figure~\ref{dp_paper}. Note that the proposed network can also be considered as a neural network (containing stack of morphological layers) where the  neurons are morphological neurons performing dilation or erosion operation.  
  
Therefore, the contribution of the paper can be summarized as follows:
\begin{itemize}
    \item We propose morphological networks that extends the basic concepts of morphological perceptron~\cite{davidson_morphology_1993} and emulates 2D dilation and 2D erosion for gray scale image processing. 
    \item Here we have utilized a pair of series of dilation and erosion operation along two different paths and intermediate outputs are combined to predict the final output. Ideally a number of paths could be incorporated where each path corresponds to a single compound morphological operator (\emph{i.e.} concatenation of dilation and erosion operators with different structuring elements). 
    \item The proposed network is evaluated for the de-raining task. We observe that a tiny morphological network (ours) is capable of producing a high quality result as that of large and complex CNNs.  
\end{itemize}
 
The rest of the paper is organized as follows. In section~\ref{sec:related} we will discuss the works related to the proposed morphological network. In section~\ref{sec:methods} we describe the building blocks of the learning structuring elements and define basic operations of mathematical morphology, \emph{i.e.}, dilation-erosion in 2D. We have evaluated our algorithm on rain dataset~\cite{fu2017clearing} and presented the results in section~\ref{sec:exp}. Lastly, we will conclude the paper in section~\ref{sec:conclude}.

\subsection{Related work}\label{sec:related} 

In our work we have used basic concepts of morphological perceptron. Morphological perceptron was introduced by \cite{davidson_morphology_1993} and the authors used morphological network to solve template identification problem. Later it was generalized by \cite{ritter_introduction_1996} to tackle the problem of binary classification by restricting the network to single layer architecture. The decision boundaries were considered as parallel to the axes. Later in \cite{sussner_morphological_1998} the network was extended to two layers and it was shown that 
the decision boundary need not be axis parallel for the classification. 

To train the structuring elements of the network, researchers~\cite{pessoa_neural_2000} have tried to use gradient descent by combining classical perceptron with morphological perceptron. In~\cite{de_a._araujo_morphological_2012} they used linear function to apply it to regression problems. With dendritic structure of morphological neurons, Zamora~\emph{et al.}~\cite{zamora_dendrite_2017} replaced the \texttt{argmax} operator by \texttt{softmax} function to overcome the problem of gradient computation and used gradient descent to train the network. 
Similar work has been done by Ranjan~\emph{et al.}~ \cite{2019arXiv190100109M} where they have learned 1D structuring elements in a dense network by back-propagation.
It may be noted that though the functions with $\max$ or $\min$ are not, in general, differentiable, those are piecewise differentiable. 
Using this property, in our work we could use gradient descent during the back-propagation to learn the 2D structuring elements for dilation and erosion operations in the network. In the next section we have defined the basic building blocks of the network, \emph{i.e.} 2D  dilation and 2D erosion by $\max$ and $\min$ operations. 

\section{Method}\label{sec:methods} 

Here we first describe the 2D dilation and erosion operations which are the basic building blocks of the network, and then discuss the network architecture in detail followed by the choice of the loss.
\begin{figure} 
\includegraphics[width=\linewidth]{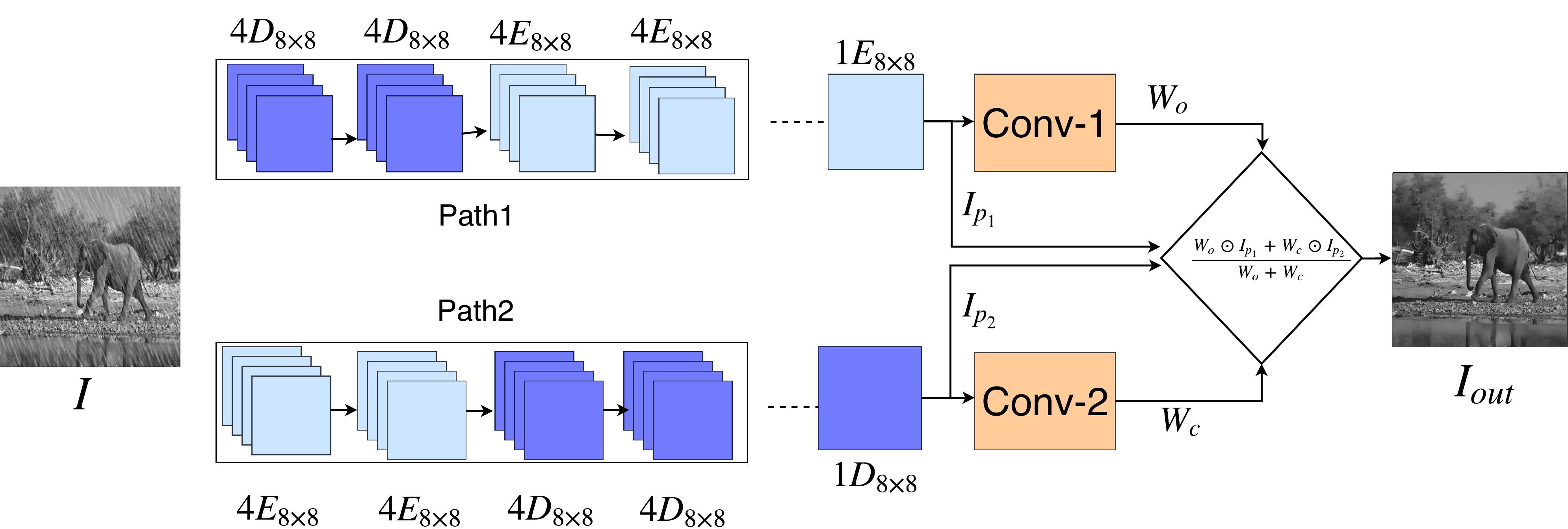}
\caption{In our proposed network (MorphoN) we consider two parallel branches of the network with an alternate sequence of dilation and erosion operators. The output of the branches are then combined with the weights predicted from the network to yield final output. Details can be found in the text.}
\label{2dDEnetwork} 
\end{figure}

\subsection{Morphological layers}\label{sec:dilero} 

The classical morphological algorithms are described using dilation and erosion operators and the morphological filtering are defined by a combination of these operators with often times utilizing different structuring elements. These operations are successfully used in many  different applications in image processing. 
In this work, we design a morphological network to emulate 
those two-dimensional gray-scale morphological operations with different structuring elements. 

Let $I$ is the input gray scale image of size $m\times n$. 
Dilation $(\oplus)$ and erosion $(\ominus)$ operation on image $I$ is defined as the following 
\begin{align}
   (I\oplus W_d)(x,y) =  \max_{i \in S_1  j \in S_2}({I(x-i,y-j)+W_d(i,j)}), \label{eq:dilation} \\
   (I\ominus W_e)(x,y) = \min_{i \in S_1  j \in S_2}({I(x+i,y+j)-W_e(i,j)}).   \label{eq:erosion} 
\end{align}
where  $W_d \in R^{a\times b}$, $W_e \in R^{a\times b}$, $S_1=\{1,2,..,a\}$  and $S_2=\{1,2,..,b\}$.
$W_d$ and $W_e$ are dilation and erosion kernels or structuring elements. After applying dilation and erosion on an image, we call the resultant as dilation map and erosion map respectively. Note that the operations defined in \eqref{eq:dilation} and \eqref{eq:erosion} 
function very similarly as the convolutional layers. The operators perform in a windowed fashion of window size $a\times b$. Padding is incorporated to have output as same size as input. The structuring 
elements ($W_d$, $W_e$) are initialized randomly and, then, optimized by back-propagation. 

In the following section we describe the network architecture using the Dilation and erosion operation on the image. 

\subsection{Morphological Network}
In morphological image processing, the \textit{Opening} and \textit{Closing} operations are commonly used as filters to remove noise. The opening operation is defined by applying dilation on an  eroded image; whereas closing operation is defined by erosion on a dilated image. For noise removal a specific ordering of opening and closing needs to be applied on the noisy image. We have considered a sequence of alternate morphological layers with  dilation and erosion to implement such filters. Each layer leads to a different dilation or erosion map because there could be different 
trained structuring elements. Multiple dilation and erosion map are useful because there could be multiple types of noise in the input image. So a single morphological network employing multiple dilation and erosion (or effectively opening and closing) would be able to filter out various types of noise. Furthermore, as shown in fig \ref{2dDEnetwork}. Here we have considered exactly two different paths of stacks of morphological layers, first starting with dilation (or closing) followed by erosion (or opening) and the second path is totally complement of the first one \emph{i.e.,} starting with erosion (or opening). 
 
 \begin{table}
 \centering
    \caption{The architectures of paths shown in Fig.~\ref{2dDEnetwork}. 4D$_{8\times8}$ denotes a layer with $4$ dilations with different trainable SEs of size $8\times8$. 4E$_{8\times8}$ is defined similarly. 2@$8\times8$--tanh denotes 2 feature map which has been produced by convolving with kernel $8\times8$ followed by tanh activation} 
  
    \begin{tabular}{c|p{0.71\linewidth}}
    \toprule
     \multicolumn{2}{c}{\textbf{Description of the MorphoN}}\\
    \hline
    {Path1-Conv} &
    4D$_{8\times8}$--4D$_{8\times8}$--4E$_{8\times8}$--4E$_{8\times8}$--4D$_{8\times8}$--4D$_{8\times8}$--4E$_{8\times8}$--4E$_{8\times8}$--4E$_{8\times8}$--2@$8\times8$--tanh--3@$8\times8$--tanh--1@$8\times8$--sigmoid  \\ 
    {Path2-Conv} &4E$_{8\times8}$--4E$_{8\times8}$--4D$_{8\times8}$--4D$_{8\times8}$--4E$_{8\times8}$--4E$_{8\times8}$--4D$_{8\times8}$--4D$_{8\times8}$--4D$_{8\times8}$--2@$8\times8$--tanh--3@$8\times8$--tanh--1@$8\times8$--sigmoid  \\ 
    \bottomrule 
    \multicolumn{2}{c}{~}\\ \toprule 
         \multicolumn{2}{c}{\textbf{Description of the smaller MorphoN}}\\
     \midrule 
        {Path1-Conv (small)} & 1D$_{8\times8}$--1D$_{8\times8}$--1D$_{8\times8}$--1E$_{8\times8}$--1E$_{8\times8}$--1D$_{8\times8}$--1D$_{8\times8}$--1E$_{8\times8}$--1E$_{8\times8}$--1E$_{8\times8}$-- 2@$8\times8$--tanh--3@$8\times8$--tanh--1@$8\times8$--sigmoid  \\ 
    {Path2-Conv (small)} &1E$_{8\times8}$--1E$_{8\times8}$--1E$_{8\times8}$--1D$_{8\times8}$--1D$_{8\times8}$--1E$_{8\times8}$--1E$_{8\times8}$--1D$_{8\times8}$--1D$_{8\times8}$--1D$_{8\times8}$--2@$8\times8$--tanh--3@$8\times8$--tanh--1@$8\times8$--sigmoid  \\ 
    \bottomrule  
    \end{tabular} 
    \label{tab:path} 
 \end{table}

 Since it is hard to know which particular path is more effective for noise removal in a specific situation, we further yield a weight map for each path to combine them to a single output. Let $W_o$, $W_c$  are the weight maps for paths starting with opening and starting with closing respectively which are of the same size as input image. We have taken \texttt{sigmoid} as activation function in the last layer so the value of each pixel in $W_o$ and $W_c$ are greater than zero and less than $1.0$. Finally, we get the output $I_{out}$ by the following equation. 
\begin{equation}
    I_{out}={{W_o \odot I_{p_{1}} + W_c \odot I_{p_{2}} } \over {W_o+W_c}}
 \end{equation}
where $I_{p_{1}}$ and $I_{p_{2}}$ are the outputs from  path1 and path2 respectively, and $\odot$ is the pixel-wise multiplication. 

\subsection{Learning of Structuring Elements} 
As defined in section \ref{sec:dilero}, dilation and erosion consist of $\max$ and $\min$ operations respectively. The expression containing $\max$ and $\min$ are piece-wise differentiable. So, we could use back propagation algorithm to learn  the structuring elements of the network as well as the weights combination. We hypothesize that SSIM~\cite{wang2004image} is a good measure which quantifies image quality degradation between two images. SSIM measure between two images $x$ and $y$ of same size is defined by
\begin{equation}
    \text{SSIM}(x,y)=\frac{(2\mu_x \mu_y +c_1)(2\sigma_{xy}+c_2) }{(\mu_x^2n+\mu_y^2+c_1)(\sigma^2+\sigma^2+c_2)}
\end{equation}
where $\mu_x$ and $\mu_y$ are the mean of the image $x$ and $y$ respectively and $\sigma_x^2$ and $\sigma_y^2$ are the variance of the image $x$ and $y$ respectively.  $\sigma_{xy}$ is covariance between $x$ and $y$. $c_1$ and $c_2$ is constant taken as $0.0001$ and $0.0009$ respectively.
To train the network we have used structural dissimilarity (DSSIM) as the objective function over a small patch of the output, where DSSIM is related to SSIM  by the following equation, 
\begin{equation}
 \text{DSSIM}(I_{out},I_{gt}) =\frac{1}{M}  \sum_{i} \frac{1-\text{SSIM}(P_{out}^i, P_{gt}^i)}{2}   
\end{equation}
where $P_{out}^{i}$ and $P_{gt}^{i}$ are $i^{th}$ spatially same patch of the network predicted output image $I_{out}$ and ground truth image $I_{gt}$  respectively. $M$ is the total number of such patches. In our experiment we have taken the patch size as $100 \times 100$. 
In practice, we combine DSSIM and MAE loss  by the following equation. 

\begin{equation}
Loss_{total} =  \text{DSSIM}(I_{out},I_{gt}) + \lambda \text{MAE}(I_{out},I_{gt})
\end{equation} 
where $\lambda$ is the weighting constant between two losses and $\text{MAE}(P_{out},P_{gt})$ is defined as follows
\begin{equation}
    \text{MAE}(I_{out},I_{gt}) = \frac{1}{N}\| I_{out}-I_{gt} \|_1 
\end{equation}
where $N$ is the number of pixels. For all the experiments we have taken $\lambda=1$. 
In the next section we evaluate the proposed morphological networks.

\section{Experiments}\label{sec:exp}

We have evaluated the proposed morphological network for image de-raining task on publicly available Rain dataset~\cite{fu2017clearing}. The dataset contains $1,000$ clean images, and for each clean image it has $14$ different rainy images with different streak orientations and sizes. Out of $1,000$, $80\%$ of the data has been considered for training, $10\%$ of the data considered for validation and remaining $10\%$ of the data \emph{i.e,} $100$ images have been kept for testing. Since all the images are of different sizes, we have resized them to $512 \times 512$ by bilinear interpolation (implementation bottleneck). Since the proposed morphological layers are designed for gray scale image, we have converted all the images to gray scale. An extension to color images is also possible by adding more channels in the morphological layers, but it is not exploited in the current work. 

In our experiment, to evaluate each path (corresponding to a single compound morphological operation), we have also trained path1 and path2 separately. Quantitative and qualitative results on path1 and path2 is also reported. 
The proposed morphological layers are implemented on Keras python scripts with back-end TensorFlow library. The evaluations have been carried out in a machine, which has a Intel Xeon 16 core processor and 128GB RAM. We have used Nvidia Titan Xp GPU for the parallel processing. 
In the next section we have  shown qualitative and quantitative evaluation on test data. 

\subsection{Parameter Settings}
For the initialization of the network, we have used the standard glorot uniform initializer. It draws samples from a uniform distribution within $[-l, l]$ where $l$ is $\sqrt(6/(f_{in}+f_{out}))$ and $f_{in}$ is the size of SE and $fan_{out}$ is the number of morphological operators acting parallel. Proposed network is concatenation of dilation and erosion layers that involves $\max$ and $\min$ operators and are piece-wise differentiable with respect to SE. Therefore, standard backpropagation method can update the SEs. In this work, the Adam optimizer is used with default parameter settings ($lr=0.001$, $\beta_1=0.9$, $\beta_2=0.999$, $\epsilon=1e-8$). \footnote{Source code : \href{https://github.com/ranjanZ/2D-Morphological-Network}{https://github.com/ranjanZ/2D-Morphological-Network}}

\subsection{Qualitative Evaluation } 
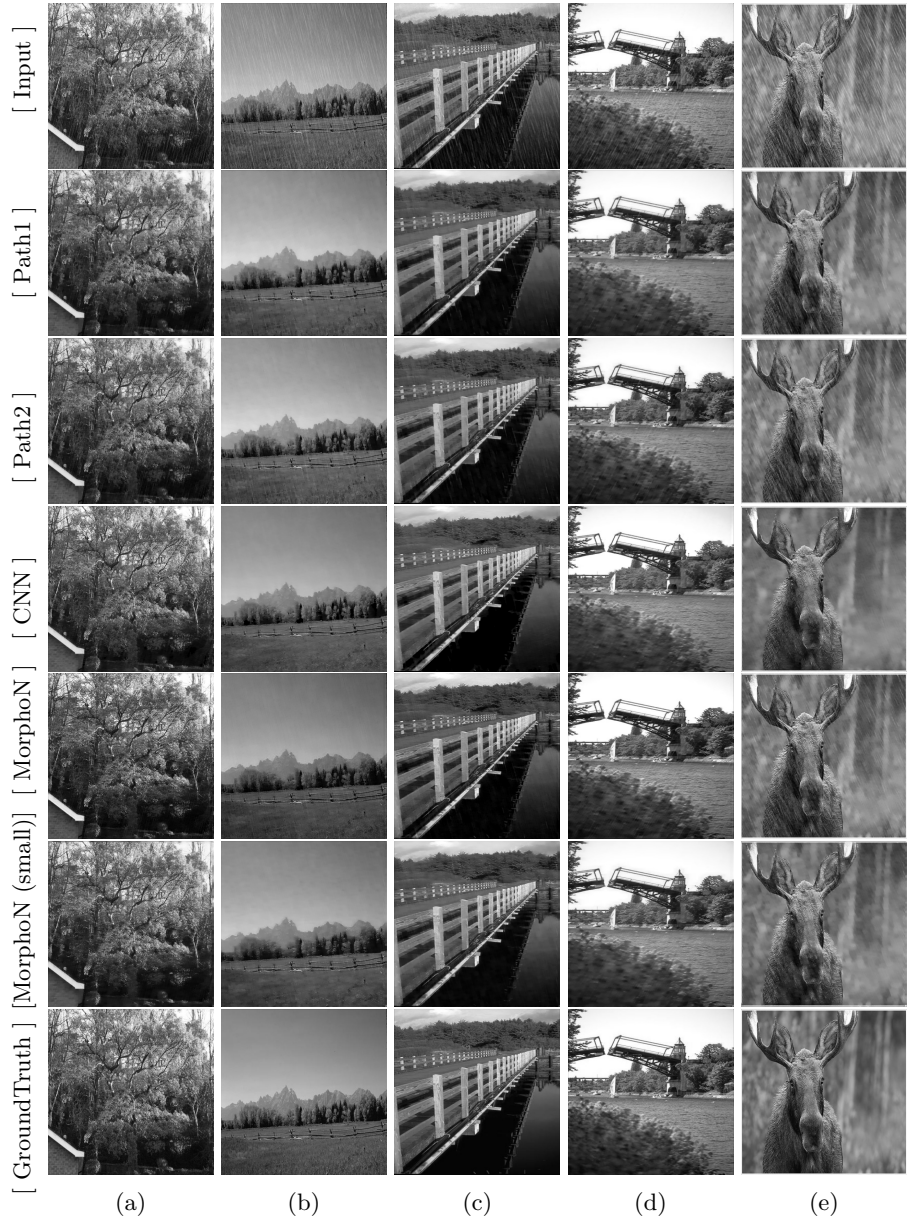
\begin{figure}
\centering
\begin{subfigure}[t]{0.03\textwidth}
\begin{picture}(1,25)
  \put(0,5){\rotatebox{90}{~~~~~[ Input ]}}
\end{picture} \\ ~\\ ~\\ ~\\  ~\\ 
\begin{picture}(1,25)
  \put(0,5){\rotatebox{90}{~~~~~[ Path1 ]}}
\end{picture} \\ ~\\ ~\\ ~\\ ~\\
\begin{picture}(1,25)
  \put(0,5){\rotatebox{90}{~~~~~[ Path2 ]}}
\end{picture} \\ ~\\ ~\\ ~\\ 
\begin{picture}(1,25)
  \put(0,5){\rotatebox{90}{~~~~~[ CNN ]}}
\end{picture} \\ ~\\ ~\\ ~\\ 
\begin{picture}(1,25)
  \put(0,5){\rotatebox{90}{~~~~~[ MorphoN ]}}
\end{picture} \\ ~\\ ~\\ ~\\  ~\\ ~\\ 
\begin{picture}(1,25)
  \put(0,5){\rotatebox{90}{~~~~~[MorphoN (small)]}}
\end{picture} \\ ~\\ ~\\ ~\\  ~\\ 
\begin{picture}(1,25)
  \put(0,5){\rotatebox{90}{~~~~~[ GroundTruth ]}}
\end{picture} 
\end{subfigure}
\begin{subfigure}[t]{0.18\textwidth}
\includegraphics[width=\textwidth]{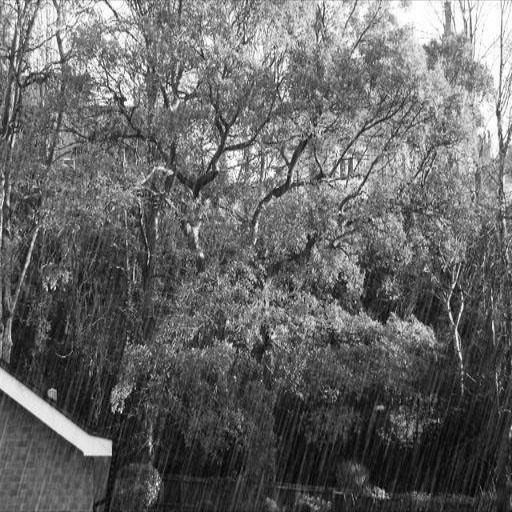}
\includegraphics[width=\textwidth]{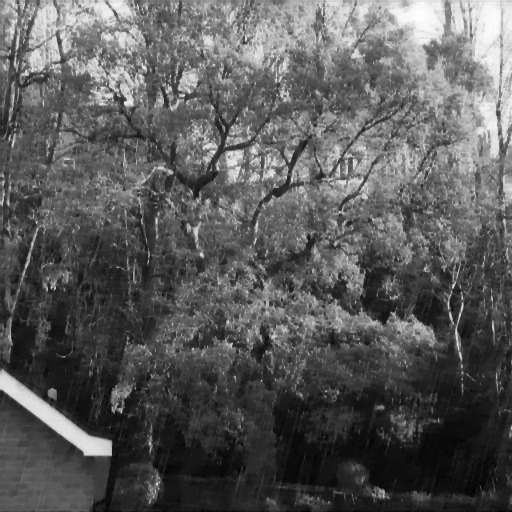}
\includegraphics[width=\textwidth]{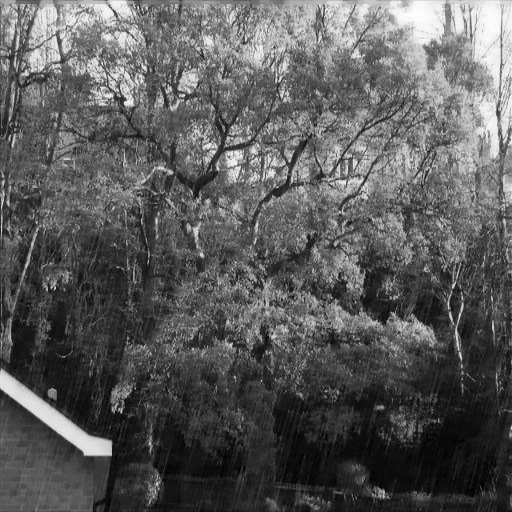}
\includegraphics[width=\textwidth]{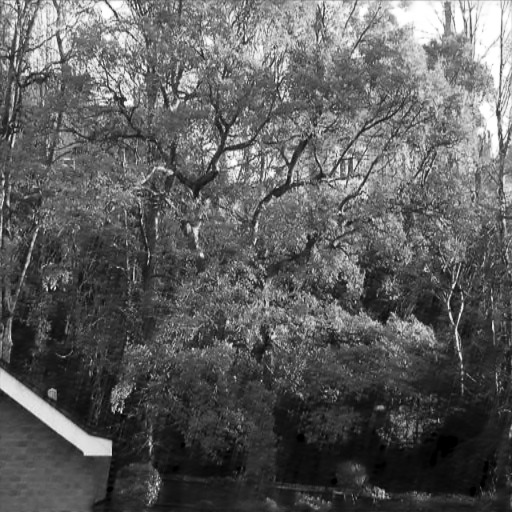}
\includegraphics[width=\textwidth]{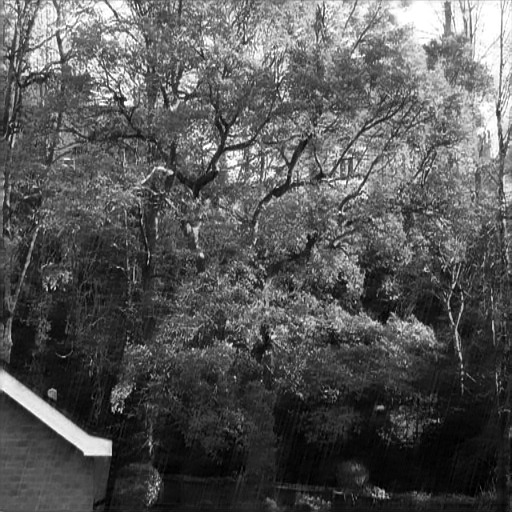}
\includegraphics[width=\textwidth]{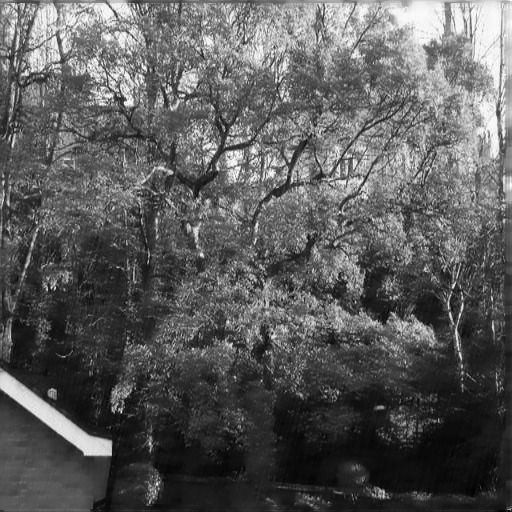}
\includegraphics[width=\textwidth]{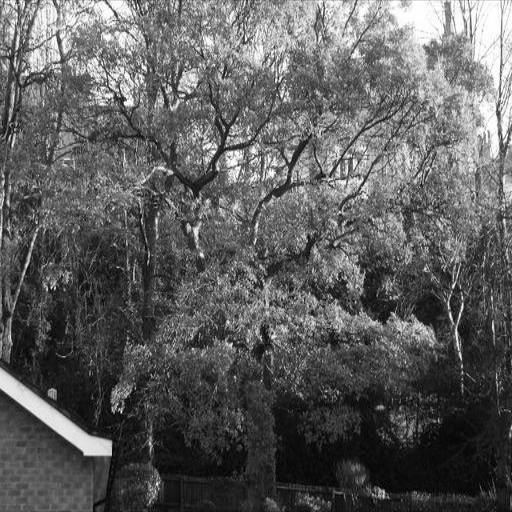}
\caption{}
\end{subfigure}
\begin{subfigure}[t]{0.18\textwidth}
\includegraphics[width=\textwidth]{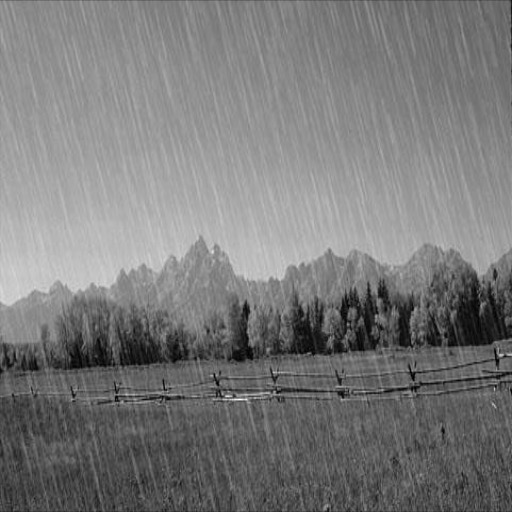}
\includegraphics[width=\textwidth]{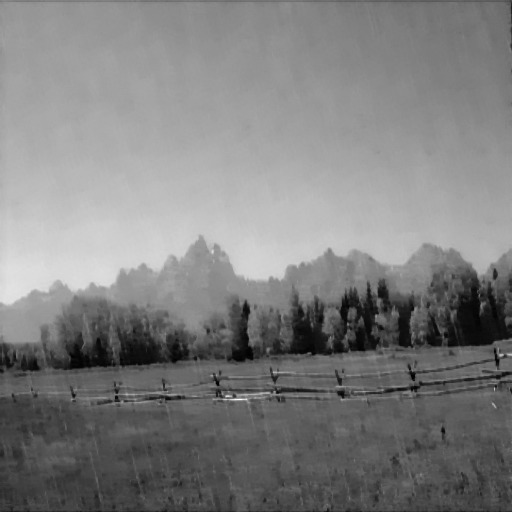}
\includegraphics[width=\textwidth]{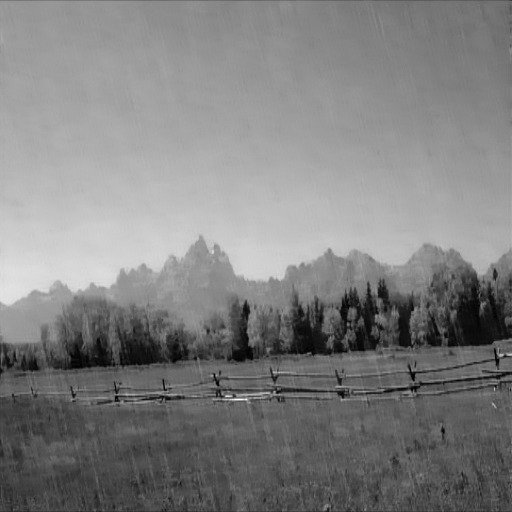}
\includegraphics[width=\textwidth]{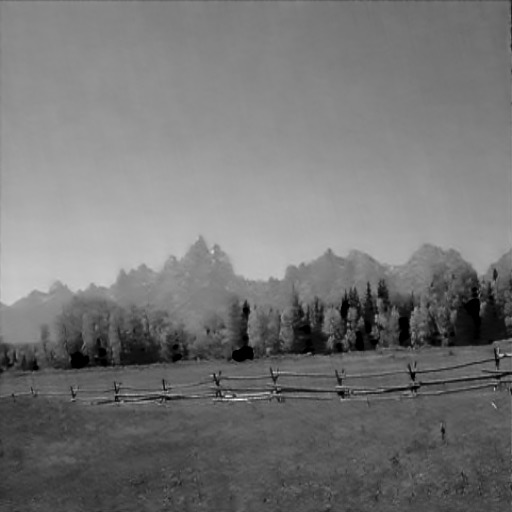}
\includegraphics[width=\textwidth]{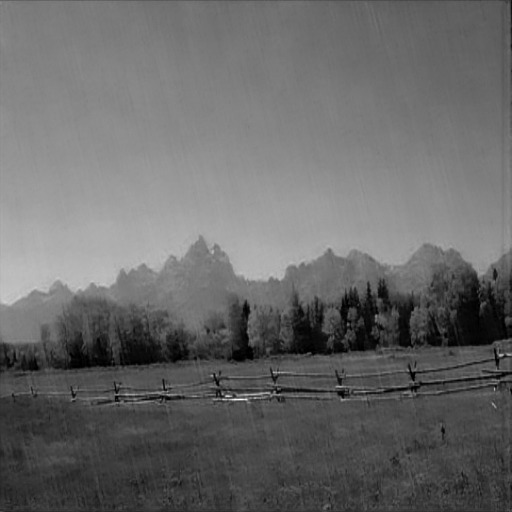}
\includegraphics[width=\textwidth]{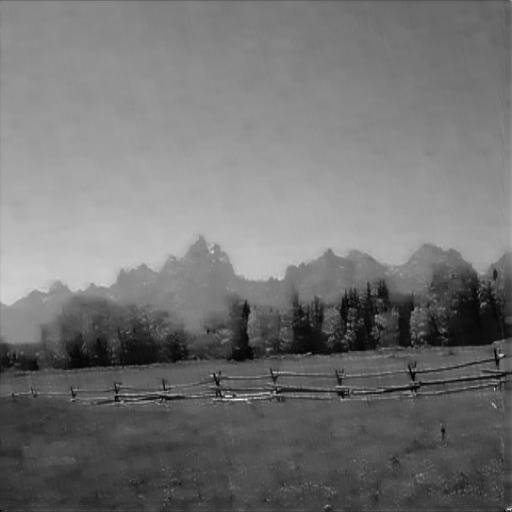}
\includegraphics[width=\textwidth]{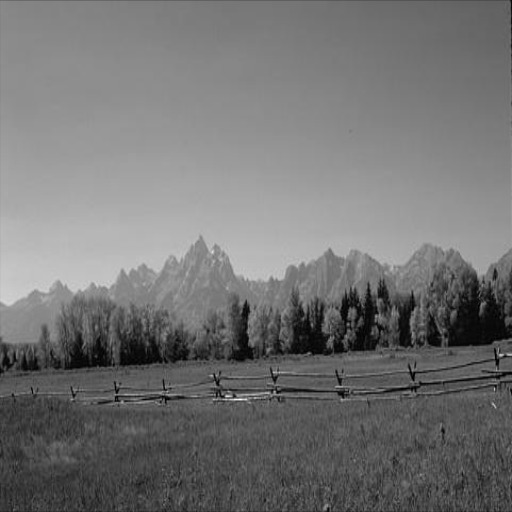}
\caption{}
\end{subfigure}
\begin{subfigure}[t]{0.18\textwidth}
\includegraphics[width=\textwidth]{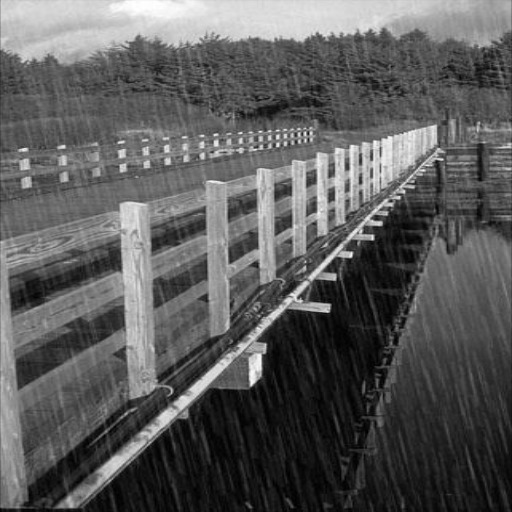}
\includegraphics[width=\textwidth]{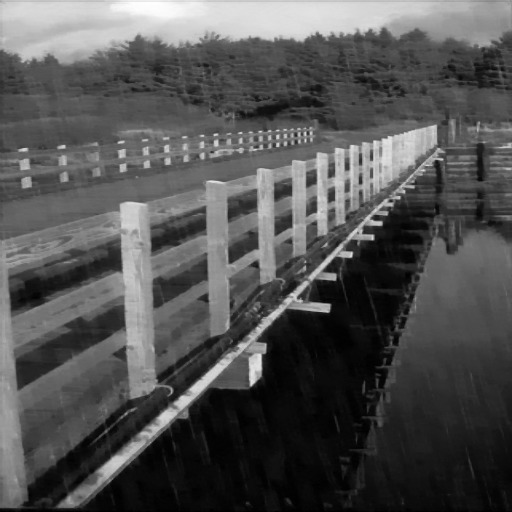}
\includegraphics[width=\textwidth]{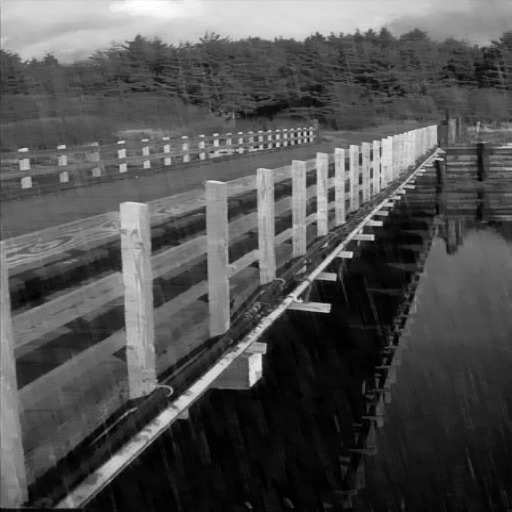}
\includegraphics[width=\textwidth]{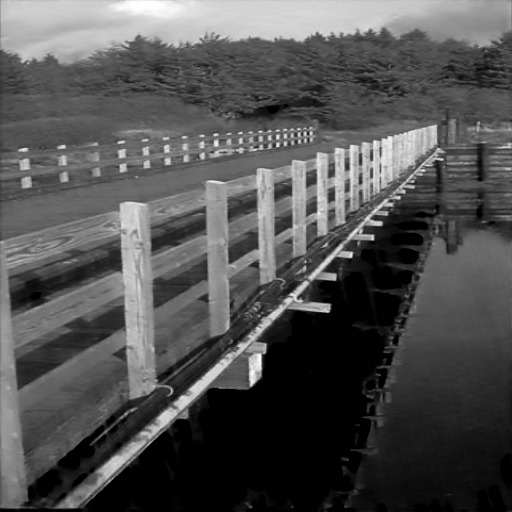}
\includegraphics[width=\textwidth]{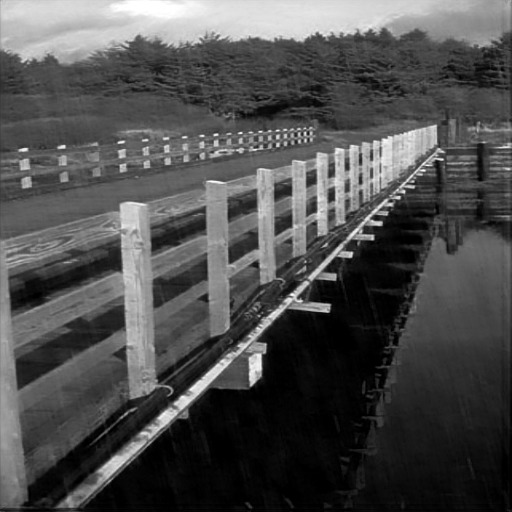}
\includegraphics[width=\textwidth]{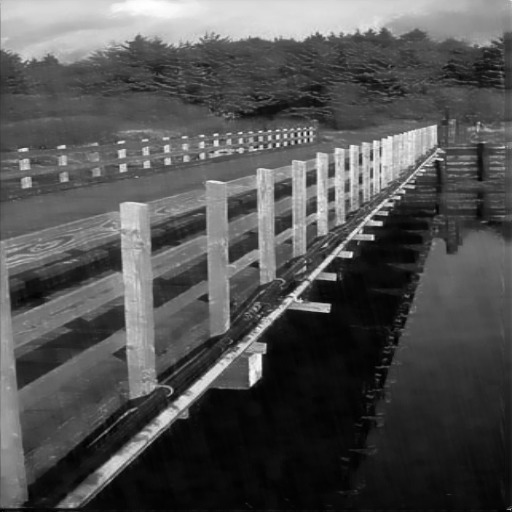}
\includegraphics[width=\textwidth]{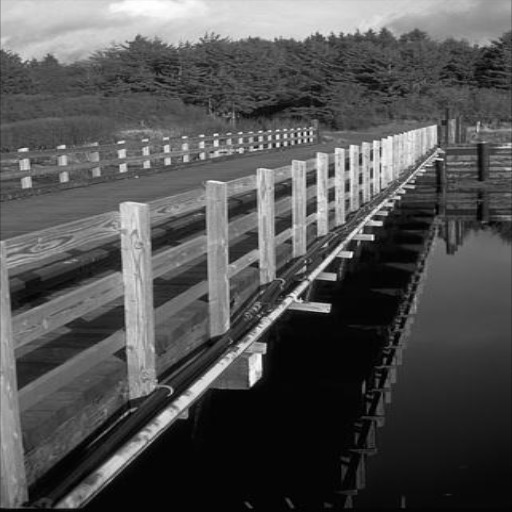}
\caption{}
\end{subfigure}
\begin{subfigure}[t]{0.18\textwidth}
\includegraphics[width=\textwidth]{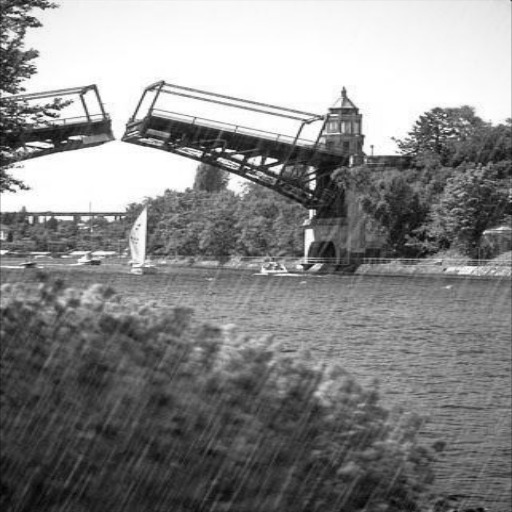}
\includegraphics[width=\textwidth]{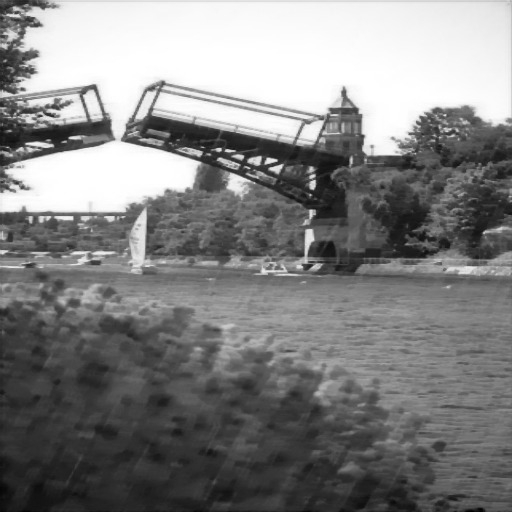}
\includegraphics[width=\textwidth]{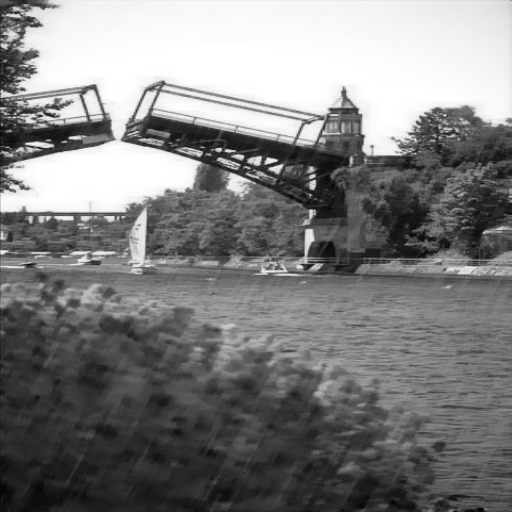}
\includegraphics[width=\textwidth]{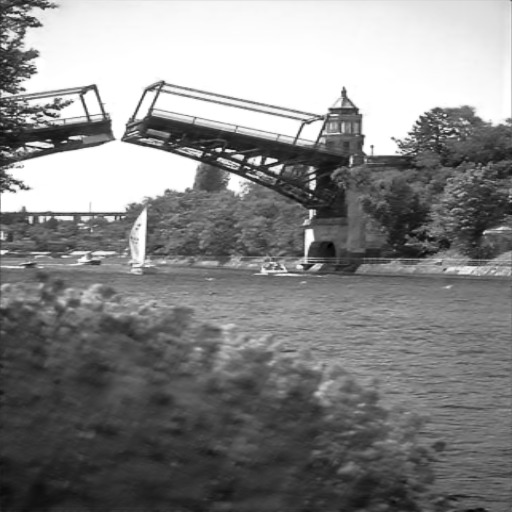}
\includegraphics[width=\textwidth]{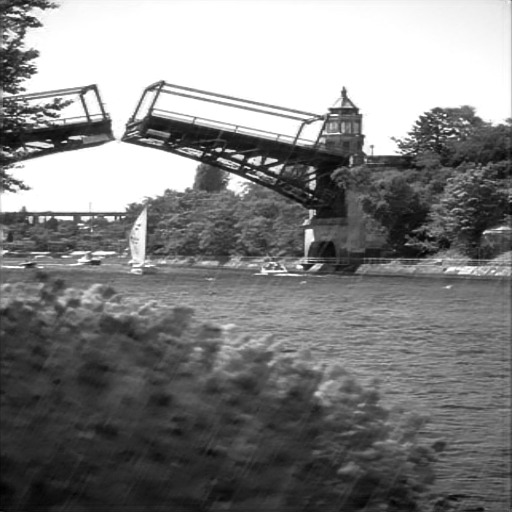}
\includegraphics[width=\textwidth]{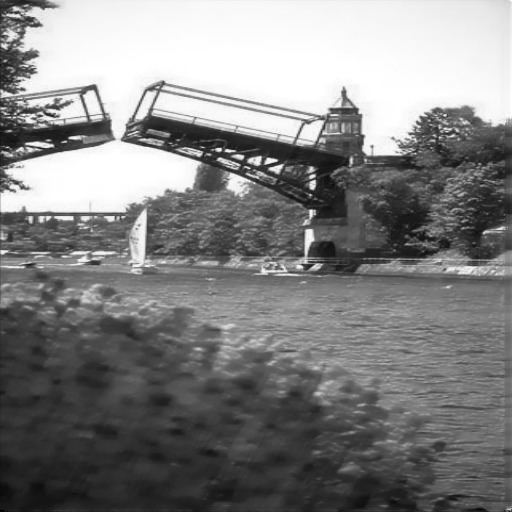}
\includegraphics[width=\textwidth]{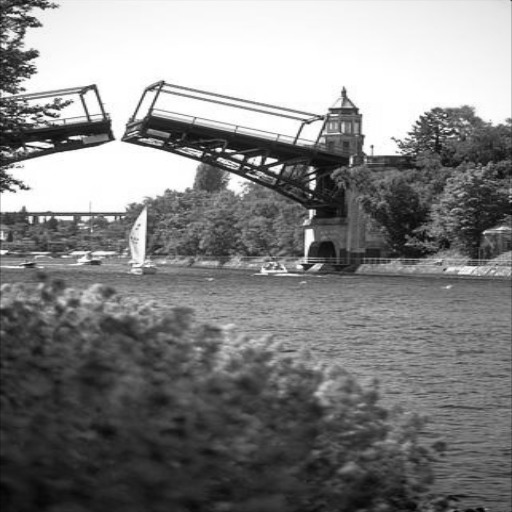}
\caption{}
\end{subfigure}
\begin{subfigure}[t]{0.18\textwidth}
\includegraphics[width=\textwidth]{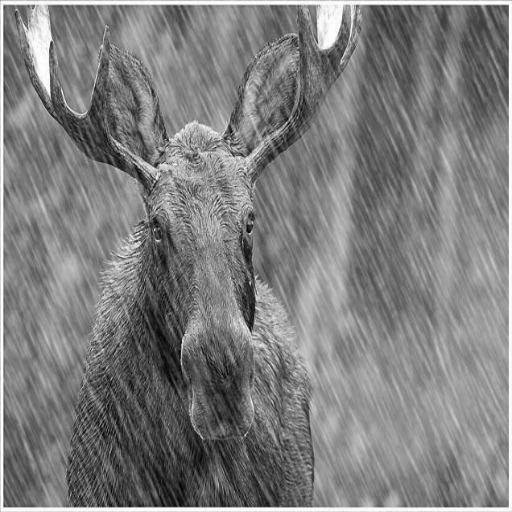}
\includegraphics[width=\textwidth]{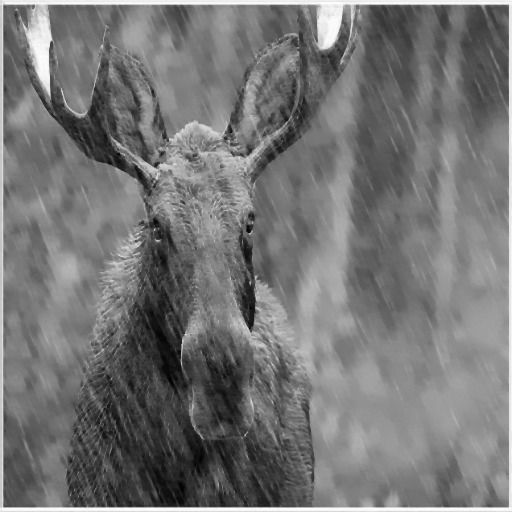}
\includegraphics[width=\textwidth]{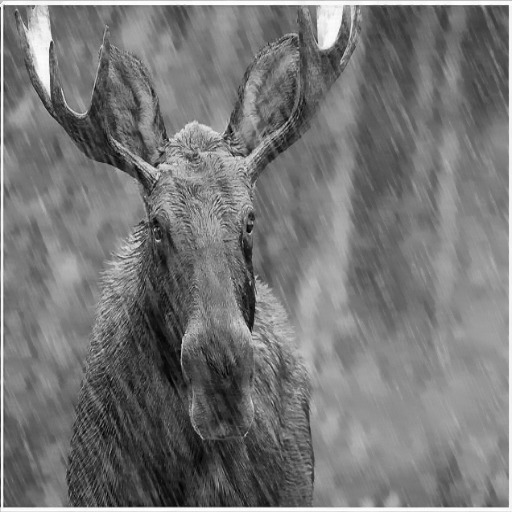}
\includegraphics[width=\textwidth]{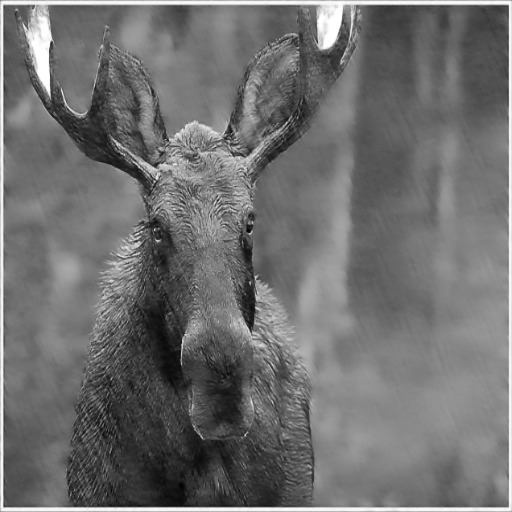}
\includegraphics[width=\textwidth]{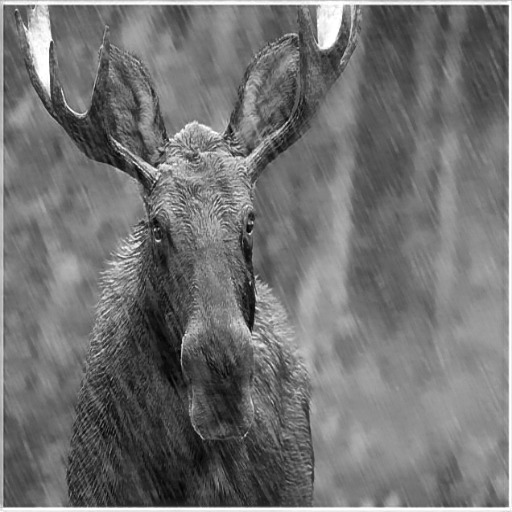}
\includegraphics[width=\textwidth]{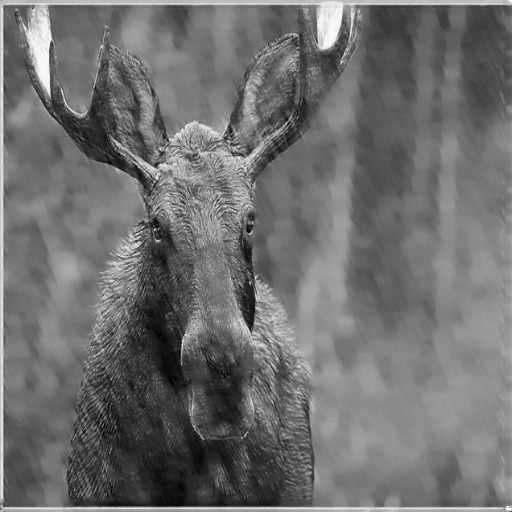}
\includegraphics[width=\textwidth]{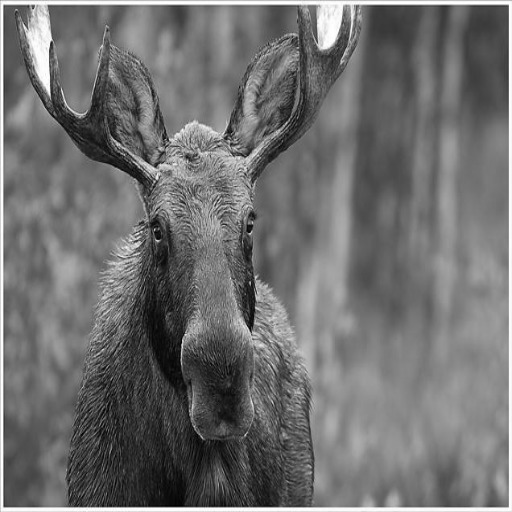}
\caption{}
\end{subfigure}

\caption{Qualitative results on Rain image dataset~\cite{fu2017clearing}. Note that the proposed morphological network produces high quality results, very similar to standard CNNs with a fraction of parameters. Note that image quality can be further improved with stacking more morphological layers and incorporating multiple paths. The architecture is not optimized for the task and the results can be further improved upon fine-tuning the network.}
\label{fig:small_network_results}
\end{figure}

We have compared our result with the results of image de-raining using a standard convolutional neural network.  To be precise, U-net architecture~\cite{ronneberger2015u} has been considered as baseline. The results of each path is also reported as baselines. In figure \ref{fig:small_network_results} we have shown the comparison of  different methods with ours. We observe that the proposed  morphological network is able to clean small rain-drops irrespective of different inclinations or rain patterns. It is interesting to see that a separately trained network along path2 produces better results compared to the network along path1. We believe that for de-raining task the noisy pixels are usually bright pixels compared to the neighbouring ones that leads to such behaviour.

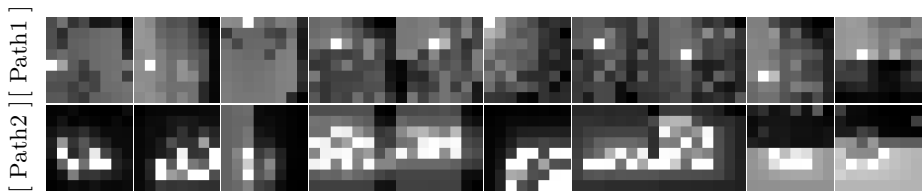
\begin{figure}
\begin{subfigure}[t]{0.03\textwidth}
\begin{picture}(1,25)
  \put(0,20){\rotatebox{90}{~~~~~[ Path1 ]}}
\end{picture} \\ 
\begin{picture}(1,25)
  \put(0,10){\rotatebox{90}{~~~~~[ Path2 ]}}
\end{picture} 
\end{subfigure}
\includegraphics[width=0.96\linewidth]{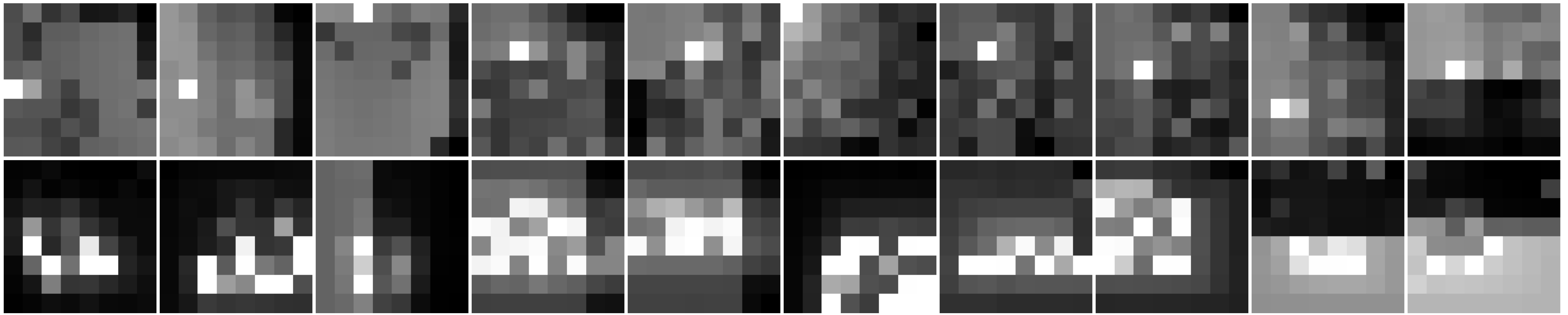}
\caption{We display the learned structuring elements at different layers of the small network along different paths. 
The structuring elements are normalized for visualization between $0$ to $255$. The most bright pixel is displayed by the maximum value of the structuring elements and the darker pixel by minimum value.}
\label{stelement}
\end{figure}

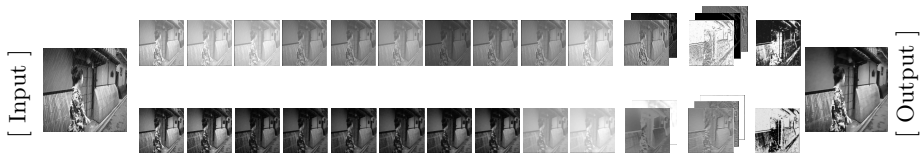
\begin{figure}
\begin{subfigure}[t]{0.03\textwidth}
\begin{picture}(1,25)
  \put(0,-10){\rotatebox{90}{~~~~~[ Input ]}}
\end{picture} 
\end{subfigure}
\includegraphics[width=0.92\linewidth]{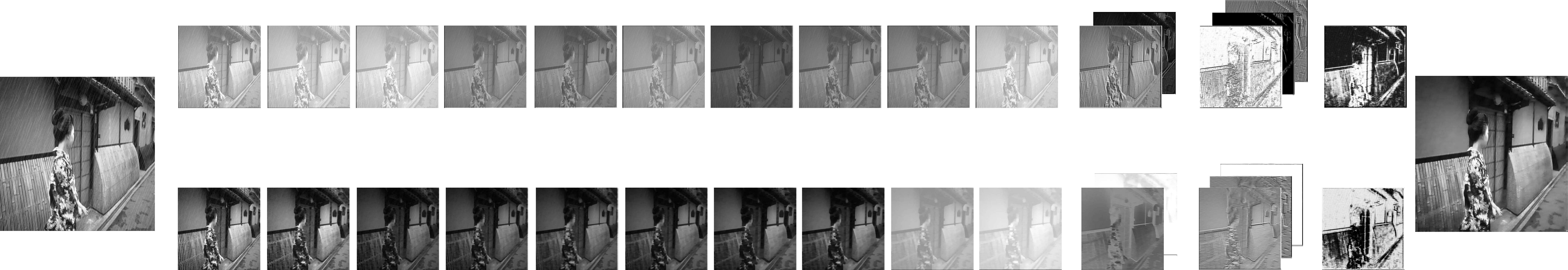}
\begin{subfigure}[t]{0.03\textwidth}
\begin{picture}(1,25)
  \put(0,-10){\rotatebox{90}{~~~~~[ Output ]}}
\end{picture} 
\end{subfigure}
\caption{Visualization of the sequential layer-wise morphological operations. The input rainy image passes through 
each layer of the network (essentially a morphological dilation / erosion operation with the trained structuring elements). The  produced filtered images are displayed at each step. The outputs of each paths are then combined by a predicted weighted combination to produce the final output.}
\label{layerout}
\end{figure}

We have also carried out our experiments taking a single dilation/erosion map instead of taking 4 dilation/erosion and term the network as MorphoN (small). The architecture is shown in table \ref{tab:path}.  Here  We have combined path1 and path2 with Conv-1  and Conv-2 respectively in the same way as we did for the original model (with 4 copies  of dilation and erosion). Using small network we are getting very similar results as the original model. In figure  \ref{stelement}, we have shown the learned structuring elements in  each path of dilation and erosion for the  small network.  We see that all the learned structuring elements are different to each other. In Fig \ref{layerout}, we have also displayed layer-wise output from the MorphoN(Small) after applying erosion and dilation. Outputs of each paths are  then  combined  by  a  predicted  weighted  combination  to produce  the  final  clear output. In Figure \ref{fig:small_network_results} we  have compared our results with CNN. 

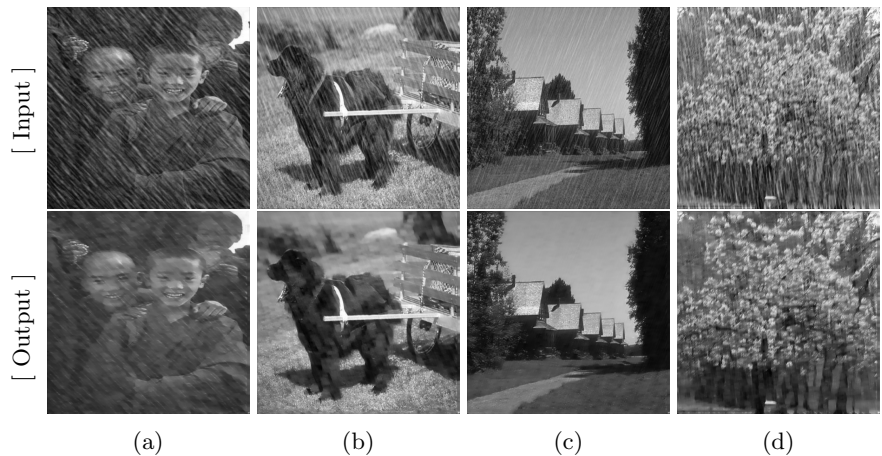
\begin{figure}
\centering
\begin{subfigure}[t]{0.03\textwidth}
\begin{picture}(1,25)
  \put(0,5){\rotatebox{90}{~~~~~[ Input ]}}
\end{picture} \\ ~\\ ~\\ ~\\  ~\\ ~\\ 
\begin{picture}(1,25)
  \put(0,0){\rotatebox{90}{~~~~~[ Output ]}}
\end{picture} 
\end{subfigure}
\begin{subfigure}[t]{0.22\textwidth}
\includegraphics[width=\textwidth]{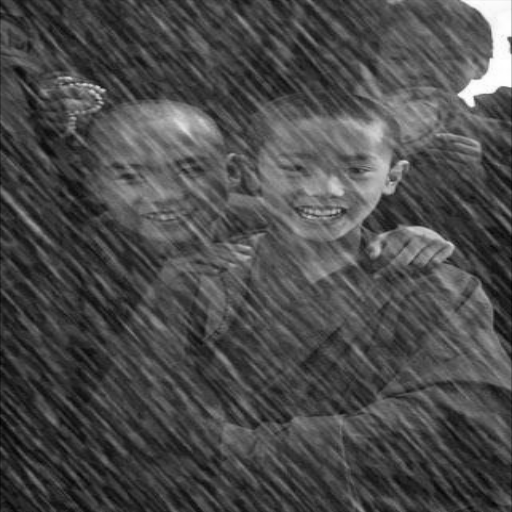}
\includegraphics[width=\textwidth]{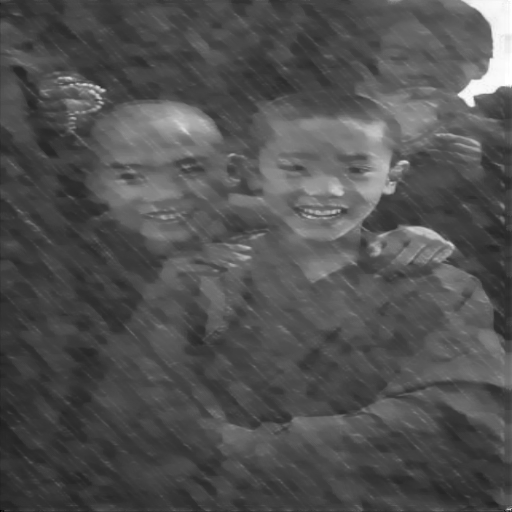}
\caption{}
\end{subfigure}
\begin{subfigure}[t]{0.22\textwidth}
\includegraphics[width=\textwidth]{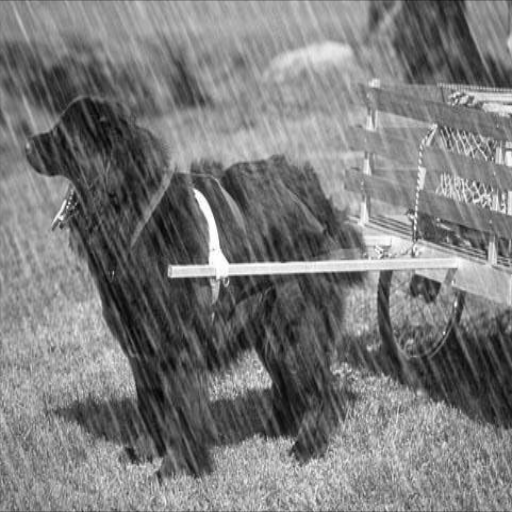}
\includegraphics[width=\textwidth]{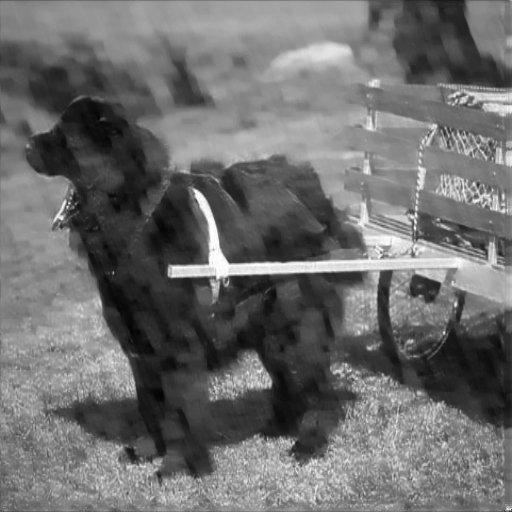}
\caption{}
\end{subfigure}
\begin{subfigure}[t]{0.22\textwidth}
\includegraphics[width=\textwidth]{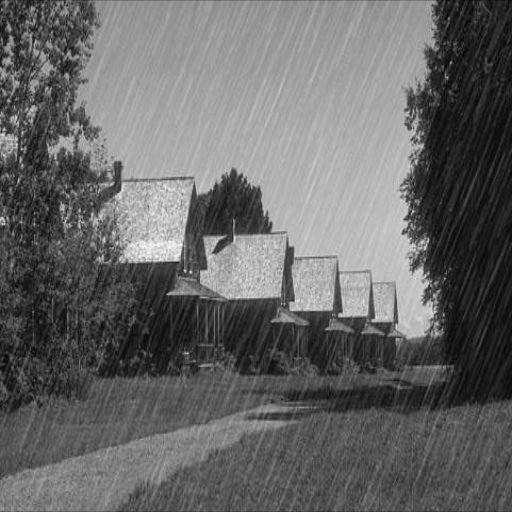}
\includegraphics[width=\textwidth]{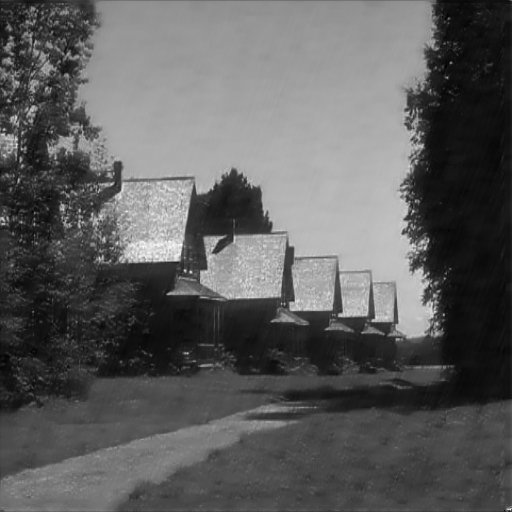}
\caption{}
\end{subfigure}
\begin{subfigure}[t]{0.22\textwidth}
\includegraphics[width=\textwidth]{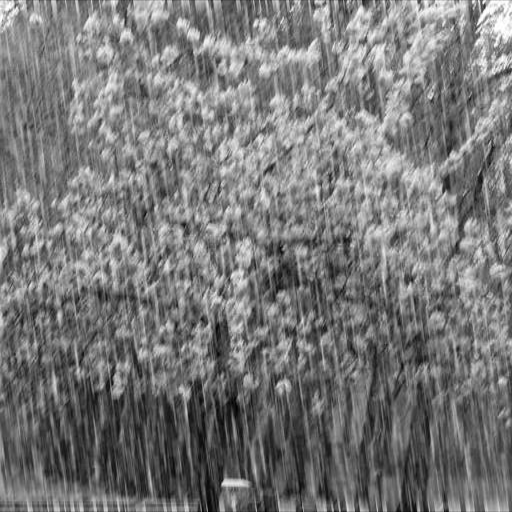}
\includegraphics[width=\textwidth]{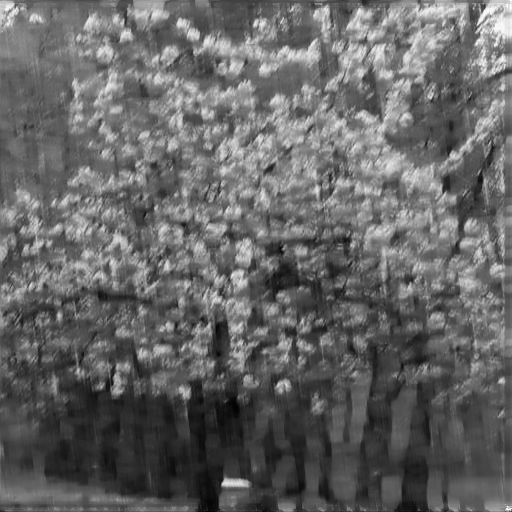}
\caption{}
\end{subfigure}
\caption{Failure cases of the proposed MorphoN (small). Note that most of the failure cases occur corresponding to the large rain structures
and we believe with larger structuring elements ($> 8\times 8$) and with more number of channels, better results can be produced. }
\label{fig:failure_cases}
\end{figure}

\begin{figure}[ht]
\centering
\begin{subfigure}[t]{0.23\textwidth}
\includegraphics[width=\textwidth]{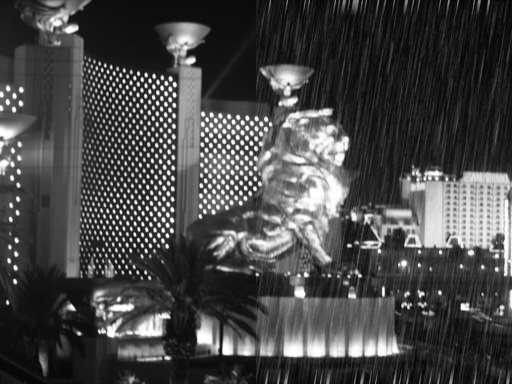}
\includegraphics[width=\textwidth]{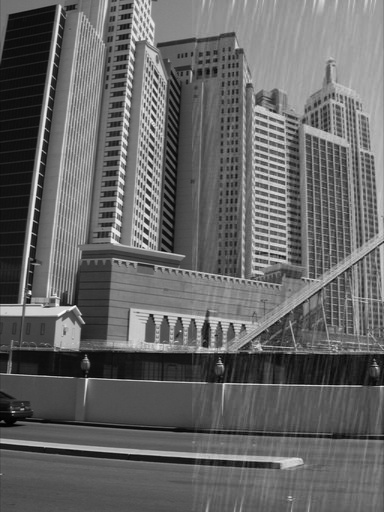}
\includegraphics[width=\textwidth]{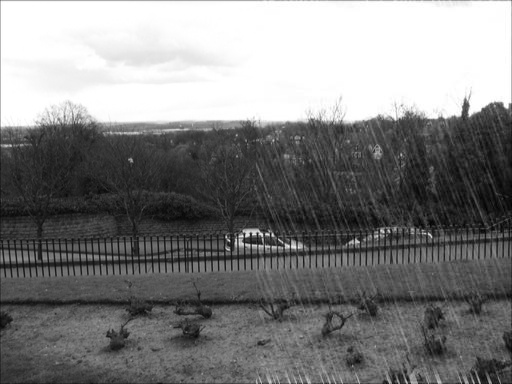}
\caption{Input}
\end{subfigure}
\begin{subfigure}[t]{0.23\textwidth}
\includegraphics[width=\textwidth]{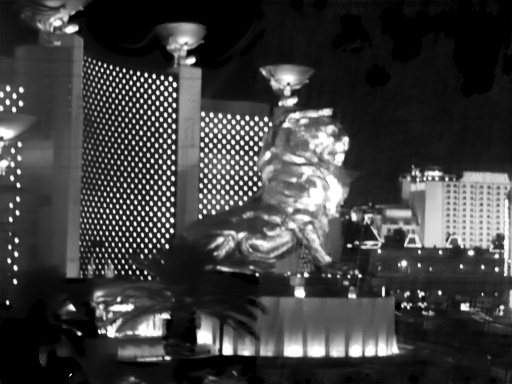}
\includegraphics[width=\textwidth]{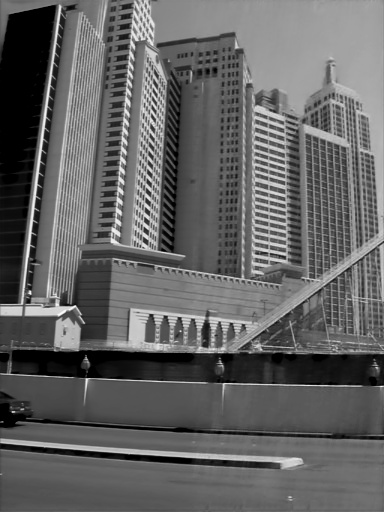}
\includegraphics[width=\textwidth]{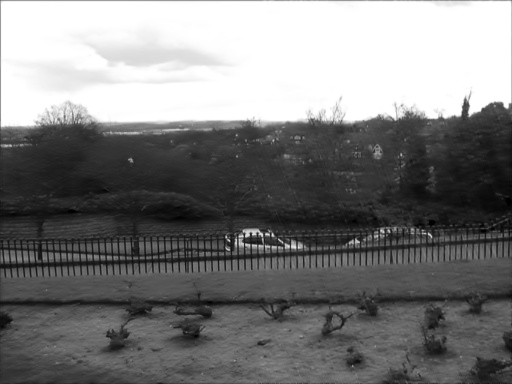}
\caption{CNN}
\end{subfigure}
\begin{subfigure}[t]{0.23\textwidth}
\includegraphics[width=\textwidth]{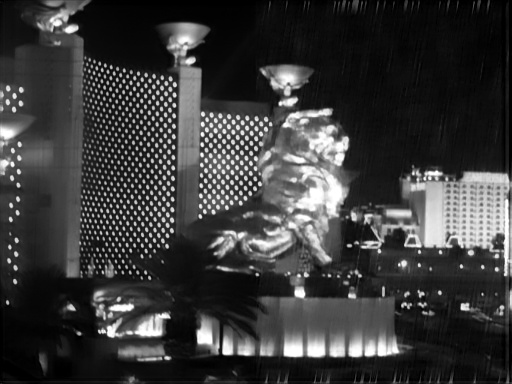}
\includegraphics[width=\textwidth]{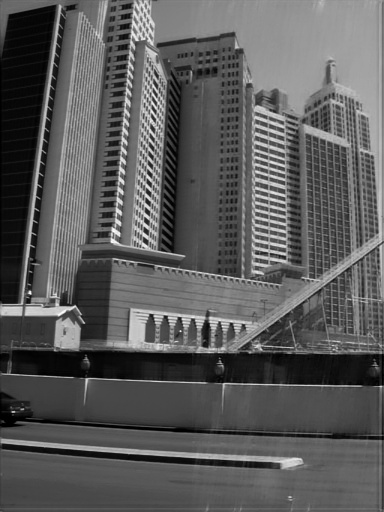}
\includegraphics[width=\textwidth]{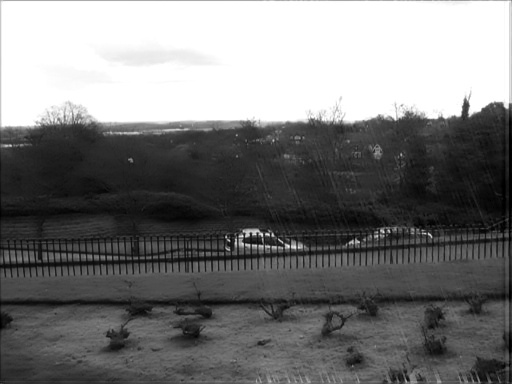}

\caption{MorphoN}
\end{subfigure}
\begin{subfigure}[t]{0.23\textwidth}
\includegraphics[width=\textwidth]{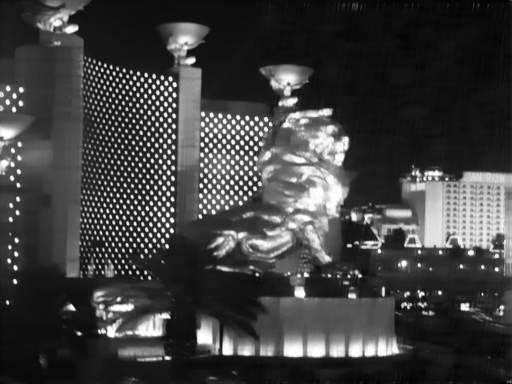}
\includegraphics[width=\textwidth]{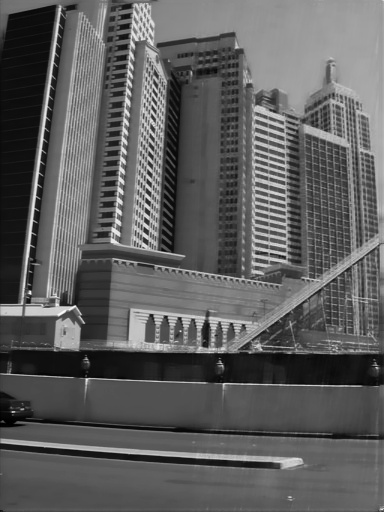}
\includegraphics[width=\textwidth]{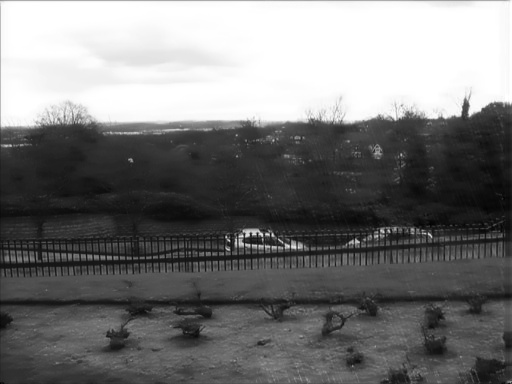}

\caption{\scalebox{0.91}{MorphoN (small)}}
\end{subfigure}
\caption{Results on partially degraded images. The partially degraded images are generated by creating rains synthetically on half of the image. Note that our method does not degrade the clean portion while removing the rain structures. }
\label{fig:partial_cases}
\end{figure}

The proposed MorphoN and MorphoN (small) produces high quality results which are very similar to standard
CNNs with a fraction  of parameters $(0.2\% - 0.04\%)$. There are some failure cases as shown in figure~\ref{fig:failure_cases}. MorphoN (small) produces blur images while removing thick rain fig \ref{fig:failure_cases}\(a,b,d\). As shown in fig \ref{fig:failure_cases}{c}, the proposed MorphoN (small) is able to clear the rain but it also clears out some structures in the house including basement. However, image  quality can be further improved  with  stacking  more  morphological  layers  and  incorporating multiple paths. The  architecture  is  not  optimized  for  the  task  and  the  results  can  be further improved upon fine-tuning the network.
A separate experiment is conducted to check if the proposed network is able to clean the noise of a partially degraded image. In Figure~\ref{fig:partial_cases}, we display such synthetic examples where half of the images were degraded by rainy structure and rest were kept unaltered. A single trained MorphoN is applied on such images--clean and the noisy part simultaneously. We observe that the clean portions are unaltered while the rainy portions are cleaned.

\begin{figure}[ht]
\centering
\begin{subfigure}[t]{0.23\textwidth}
\includegraphics[width=\textwidth]{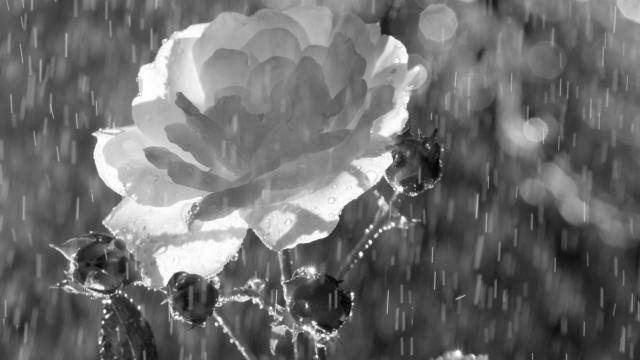}
\includegraphics[width=\textwidth]{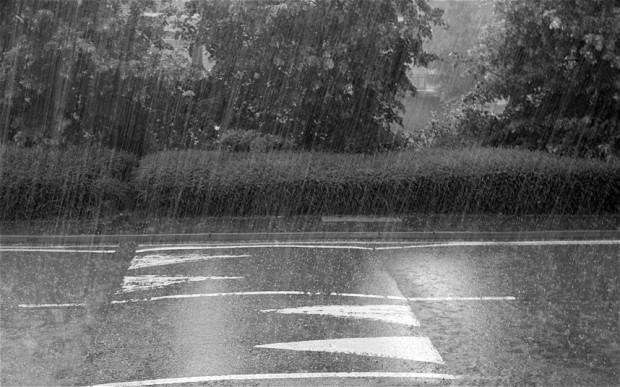}
\includegraphics[width=\textwidth]{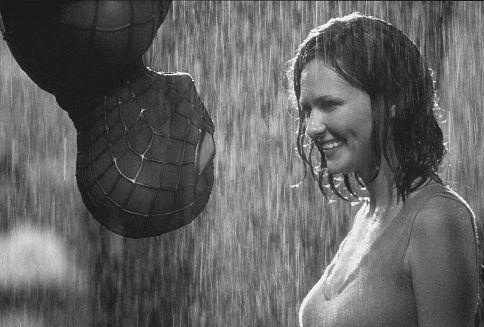}
\caption{Input}
\end{subfigure}
\begin{subfigure}[t]{0.23\textwidth}
\includegraphics[width=\textwidth]{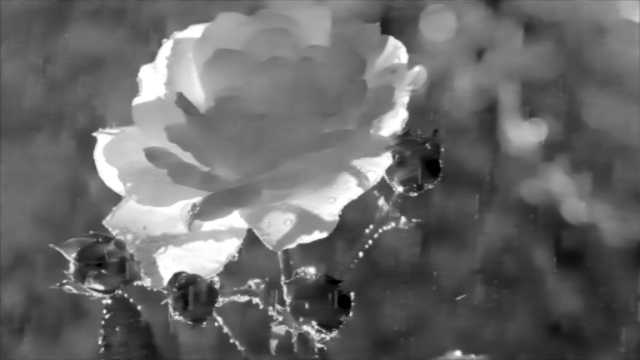}
\includegraphics[width=\textwidth]{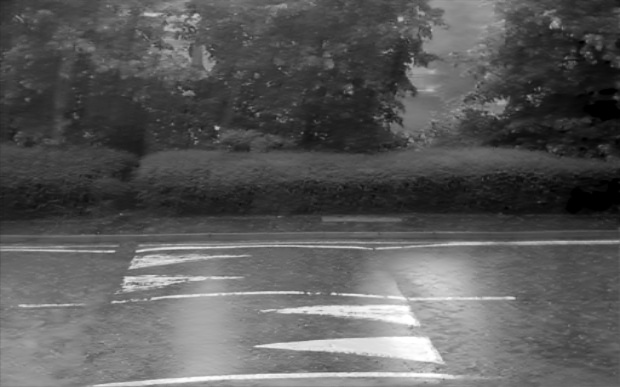}
\includegraphics[width=\textwidth]{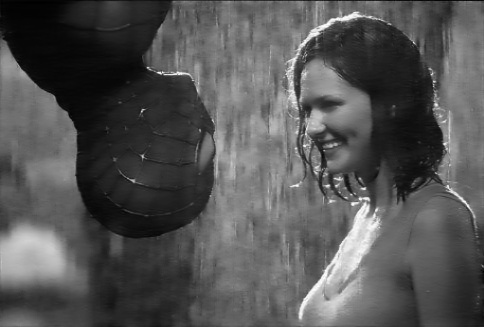}
\caption{CNN}
\end{subfigure}
\begin{subfigure}[t]{0.23\textwidth}
\includegraphics[width=\textwidth]{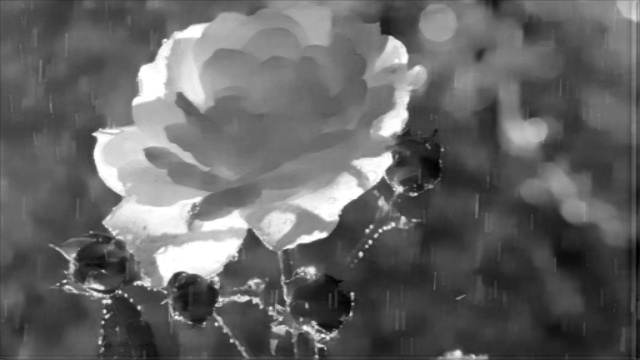}
\includegraphics[width=\textwidth]{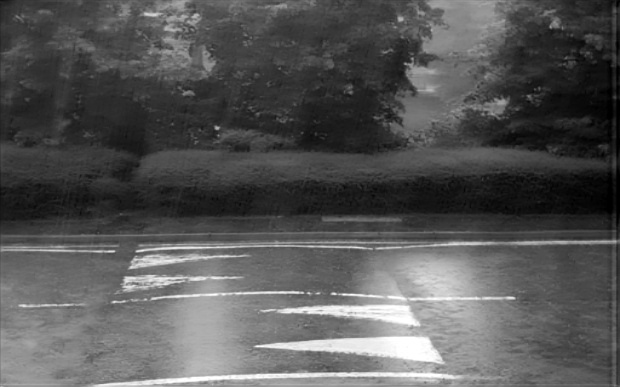}
\includegraphics[width=\textwidth]{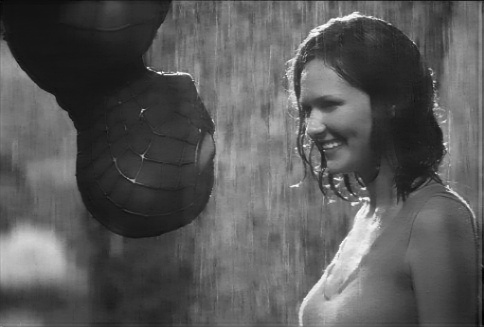}

\caption{MorphoN}
\end{subfigure}
\begin{subfigure}[t]{0.23\textwidth}
\includegraphics[width=\textwidth]{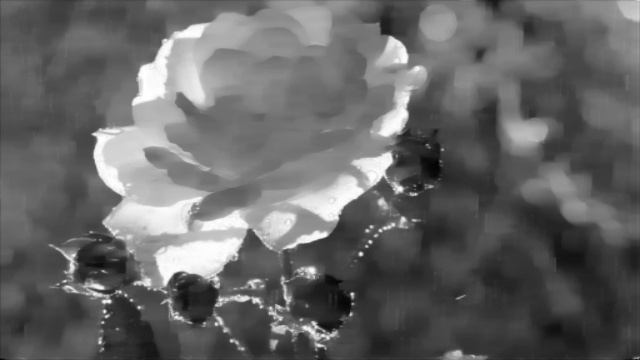}
\includegraphics[width=\textwidth]{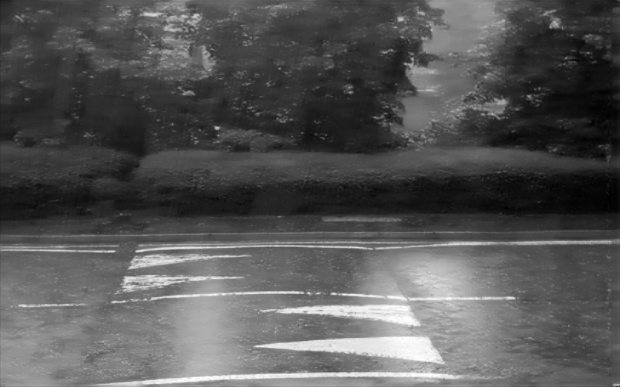}
\includegraphics[width=\textwidth]{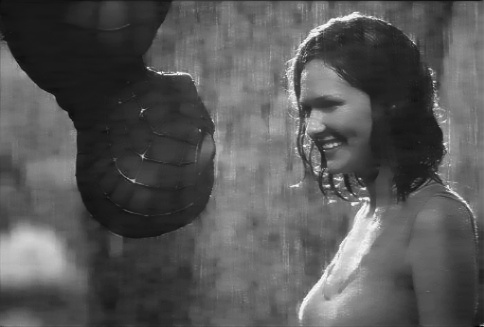}

\caption{\scalebox{0.91}{MorphoN (small)}}
\end{subfigure}
\caption{Real world examples: The proposed method and the baselines are evaluated on a number of real world examples. Our much smaller network produces results on par with the baselines. }
\label{fig:real_world}
\end{figure}

\subsection{Quantitative Evaluation}
For quantitative evaluation, we have evaluated our algorithm in terms of SSIM~\cite{wang2004image} and PSNR values. The estimated de-rain 
image is compared against the ground truth clean image. The methods are applied to all the images of the test dataset and the average value is reported. In table~\ref{quant}, we have reported the results of different methods on test data of Rainy dataset. CNN (U-Net) archives  SSIM and PNSR on an average about $0.92$ and $29.12$ respectively, whereas our network gives similar results, \emph{i.e.,}  $0.92$ and $28.03$. Our MorphoN (small) network  also produces similar results to MorphoN. We have also reported the number of parameters in the Table~\ref{quant}. Notice that MorphoN (small) with $0.04\%$ numbers of parameters of CNN  produces similar results with CNN.


\subsection{Real Data}
A dataset with real images is collected by capturing photographs of rain at different outdoor scenes and during different times of the day. Some sample images and evaluation results are displayed in Figure~\ref{fig:real_world}. Note that as the ground-truth clean images are unavailable, we only consider qualitative comparison for the evaluation. The MorphoN, trained with synthetic data,  consistently produces similar or better results than the baselines. 

\section{Conclusion}\label{sec:conclude}
    In this work, a morphological network is proposed that emulates classical morphological filtering \emph{i.e.,} 2D dilation and 2D erosion for gray scale image processing. The proposed network architecture consists of a pair of sequences of morphological layers with exactly two different paths where the outputs are combined to predict the final result. 
    We evaluated the proposed network for the de-raining task and obtained very similar results with heavy-weighted 
    convolutional neural networks. The proposed network is not tailored for de-raining and could be applied to any other filtering task. Further, this is one of the forerunner work and it opens a many directions of future research, for example, best architecture search for morphological networks. The source code is shared to encourage reproducibility and facilitate .

\begin{table}\setlength{\tabcolsep}{0.8em} 
    \centering
    \caption{Results achieved on the rain dataset~\cite{fu2017clearing} by different networks. Note that with a tiny morphological network compared to a standard CNN (U-Net)~\cite{ronneberger2015u}, a similar accuracy can be achieved. }
    \label{tab:circle_comp}
    \scalebox{0.87}{
    \begin{tabular}{c|c|c|c|c|c|c}
    \toprule 
    \textbf{Metric} & Input &CNN & Path1 & Path2 & MorphoN&\scalebox{0.95}{MorphoN (small)}  \\ 
    \hline 
     $\#$Parameters&-&6,110,773&7,680 & 7,680&16,780 &2,700\\
     \scalebox{0.95}{$\#$Params w.r.t. CNN} &-&100.0$\%$&0.12$\%$&0.12$\%$&0.27$\%$&\textbf{0.04$\%$} \\
     SSIM & 0.85&\textbf{0.92} & 0.87 &0.90& \textbf{0.92}&0.91 \\
     PSNR & 24.3&\textbf{29.12} & 26.27 & 27.20 &28.03 &27.45  \\
    \bottomrule 
    \end{tabular}
    }
    \label{quant}
\end{table}

\newpage

\section{Acknowledgements}
Initial part of the experiment has been carried out on Intel AI DevCloud. Authors want to acknowledge Intel for that.
\bibliographystyle{splncs04}
\bibliography{ref}

\end{document}